\documentclass{article}
\usepackage{ijcai23}

\usepackage[switch]{lineno}
\usepackage{times}
\usepackage{soul}
\usepackage{url}
\usepackage[hidelinks]{hyperref}
\usepackage[utf8]{inputenc}
\usepackage[small]{caption}
\usepackage{graphicx}
\usepackage{amsthm}
\usepackage{booktabs}
\usepackage{algorithm}
\usepackage{algorithmic}
\usepackage{amsmath}

\usepackage{tabularx}
\usepackage{multirow}
\usepackage{amsfonts}

\title{SS-BSN: Attentive Blind-Spot Network \\ for Self-Supervised Denoising with Nonlocal Self-Similarity}

\author{
Young-Joo Han$^{1,2}$
\And
Ha-Jin Yu $^{1}$\footnote{Corresponding Author}\
\affiliations
$^1$School of Computer Science, University of Seoul\\
$^2$Advanced Technology R\&D Center, Vieworks
\emails
orora71@gmail.com,
hjyu@uos.ac.kr
}

\begin{document}
\maketitle

\begin{abstract}
Recently, numerous studies have been conducted on supervised learning-based image denoising methods. However, these methods rely on large-scale noisy-clean image pairs, which are difficult to obtain in practice. Denoising methods with self-supervised training that can be trained with only noisy images have been proposed to address the limitation. These methods are based on the convolutional neural network (CNN) and have shown promising performance. However, CNN-based methods do not consider using nonlocal self-similarities essential in the traditional method, which can cause performance limitations. This paper presents self-similarity attention (SS-Attention), a novel self-attention module that can capture nonlocal self-similarities to solve the problem. We focus on designing a lightweight self-attention module in a pixel-wise manner, which is nearly impossible to implement using the classic self-attention module due to the quadratically increasing complexity with spatial resolution. Furthermore, we integrate SS-Attention into the blind-spot network called self-similarity-based blind-spot network (SS-BSN). We conduct the experiments on real-world image denoising tasks. The proposed method quantitatively and qualitatively outperforms state-of-the-art methods in self-supervised denoising on the Smartphone Image Denoising Dataset (SIDD) and Darmstadt Noise Dataset (DND) benchmark datasets. 
%
\end{abstract}

\section{Introduction} 
Image denoising is the process of recovering clean images from noisy images and plays an essential role in various computer vision tasks. It is an inverse ill-posed problem, which means that images should be restored from noisy images that may have numerous arbitrary noises. Especially, image denoising is indispensable when it is inevitable to obtain noisy images due to hardware limitations or healthcare issues, such as astronomical imaging or medical imaging.
Recently, numerous studies have been conducted on deep learning-based denoising methods to solve the image denoising problem. As in other deep learning-based image processing fields, studies on image denoising methods with supervised training were conducted first~\cite{DnCNN,DANet,Restormer,SwinIR}. Supervised denoising methods have shown superior performance compared to traditional methods~\cite{BM3D,NLM}. However, these methods rely on large-scale noisy-clean image pairs that are difficult to obtain in practice. For example, in the medical imaging field, obtaining large-scale noisy-clean image pairs is almost impossible. 

\begin{figure}[t]
    \renewcommand{\wp}{\linewidth}
    \centering
    \includegraphics[width=0.90\linewidth]{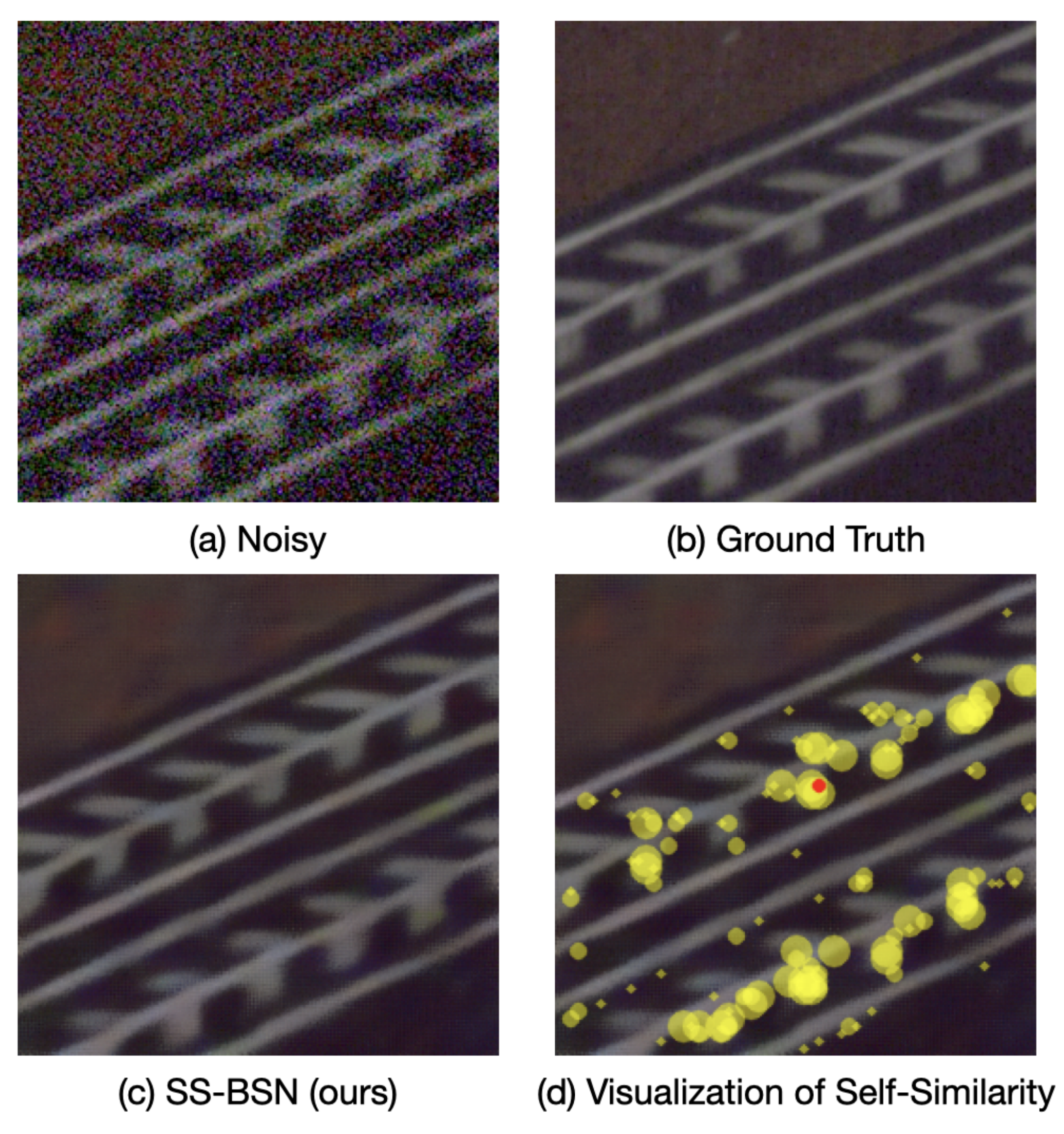}
    \\
    \vspace{-0mm}
    \caption{
        \textbf{Visualization of our proposed method on the SIDD dataset.} (a) Noisy input image, (b) ground truth image, (c) denoised estimates of our method, and (d) visualization of self-similarity. In (d), the yellow circles show the pixels with high similarity with the masked pixel (red circle) predicted by our method. A larger radius of the yellow circles indicates high similarity.
        }
    \label{fig:fig1}
    \vspace{-0mm}
\end{figure}

The Noise2Noise~\cite{N2N} method was proposed to alleviate the data collection problem. This method has proved that denoising neural networks can be trained with noisy-noisy image pairs. However, collecting numerous noisy-noisy image pairs is only possible in limited environments; thus, the difficulty of collecting the data remains.
Denoising methods with self-supervised training that can be trained with only noisy images have been proposed to address this problem~\cite{N2V,N2S,S2S,APBSN}. These methods demonstrate how to learn denoising neural networks with only noisy images using the blind-spot strategy. The blind-spot strategy avoids identity mapping by learning to predict artificially missing pixels using adjacent pixels. Therefore, denoising neural networks can be trained with only noisy images (i.e., without pairs of images). 
Based on this strategy, denoising methods with self-supervised training have been actively studied. In particular, the recently proposed AP-BSN~\cite{APBSN} has shown promising performance in real-world denoising tasks using asymmetric pixel-shuffle downsampling in the training and testing phases. However, performance degradation still occurs compared to the denoising methods using supervised or weakly supervised methods.
Essentially, the noise that needs to be eliminated in image denoising is subject to statistical fluctuation~\cite{N2Sim}. Therefore, in the early research on image denoising, studies focused on determining similar nonlocal patches and generating denoised estimates by averaging the patches~\cite{NLM,BM3D}. These studies have shown promising performance even though the studies are non-learning-based methods. Figure 1 visualizes nonlocal self-similarities in an image. However, unlike traditional methods, recent studies based on convolutional neural networks (CNNs) do not give much consideration to obtaining information from nonlocal self-similarities because the convolutional operation used in the CNN is based on local connectivity. This characteristic of the CNN can cause performance limitations in image denoising.
Recently, the transformer model with self-attention~\cite{Transformer} has achieved great success in various areas (e.g., natural language processing and high-level vision). One of the advantages of self-attention in the transformer-based model compared to the existing CNN-based model is the long-range dependency that reflects global information. The patch embedding method is adopted to apply self-attention to high-level vision tasks. For instance, the standard vision transformer (ViT)~\cite{ViT} model directly splits the image into ${16\times16}$ nonoverlapping patches. This approach can enable applying a transformer-based model in high-level vision tasks, which increases the complexity quadratically with spatial resolution. However, unlike high-level vision tasks, low-level vision tasks, such as denoising, are performed in a pixel-wise manner. Therefore, due to computational complexity, it is almost impossible to apply the patch embedding method to adopt self-attention in low-level vision tasks.
Despite the shortcoming, there have been a few efforts to apply the notion of self-attention in supervised image denoising. However, these methods calculate self-attention within a limited window size in a pixel-wise manner~\cite{SwinIR,IPT} or calculate self-attention in a channel-wise manner~\cite{Restormer}. Therefore, it is difficult to say that these methods sufficiently achieve the advantage of long-range dependency by using self-attention in a pixel-wise manner.
In this paper, we propose a simple and intuitive pixel-wise self-attention module called self-similarity-based self-attention (SS-Attention). Furthermore, we integrate SS-Attention into the blind-spot network called self-similarity-based blind-spot network (SS-BSN), which can be trained in a self-supervised manner. Unlike the previous self-attention module in image denoising, SS-Attention focuses on capturing the long-range dependency of a self-attention mechanism and obtaining information from nonlocal self-similarities that are overlooked by the existing CNN-based denoising methods.
As mentioned, it is infeasible to apply the self-attention mechanism of the classic vision transformer to denoising neural networks in a pixel-wise manner, because the complexity of the self-attention mechanism increases quadratically with spatial resolution. To solve this problem, we designed the lightweight self-attention module by removing or simplifying the components (e.g., linear transforms) of the existing classic self-attention module. To further simplify the self-attention module, we adopt grid attention~\cite{MaxViT}. Grid attention is indispensable due to the architectural characteristics of the dilated blind-spot network (D-BSN)~\cite{DBSN} which the proposed blind-spot network is based on. We also provide a hyperparameter that can control sparsity. By controlling sparsity, users can control the size of the attention map, which determines the complexity of the self-attention module. This simplified self-attention module is not expected to represent semantic information well compared to the existing classic self-attention module. However, it is enough to achieve nonlocal self-similarities in an image which we focus on.
Our contributions are as follows: we propose SS-Attention, a simple and intuitive self-attention module that focuses on the long-range dependency of a self-attention mechanism to obtain useful information from nonlocal self-similarities in an image. Additionally, we propose SS-BSN, a blind-spot network with SS-Attention that can be trained in a self-supervised manner for image denoising. Specifically, our SS-BSN is designed to effectively capture nonlocal self-similarities by using denoised features. To verify our model, we compared real-world denoising performance with various competitive baselines with the Smartphone Image Denoising Dataset (SIDD)~\cite{SIDD} and Darmstadt Noise Dataset (DND)~\cite{DND} datasets. The experiments demonstrate that the model outperforms other baselines that can be trained in a self-supervised manner. 

\begin{figure*}[t]
    \renewcommand{\wp}{0.80\linewidth}
    \centering
    \includegraphics[width=\linewidth]{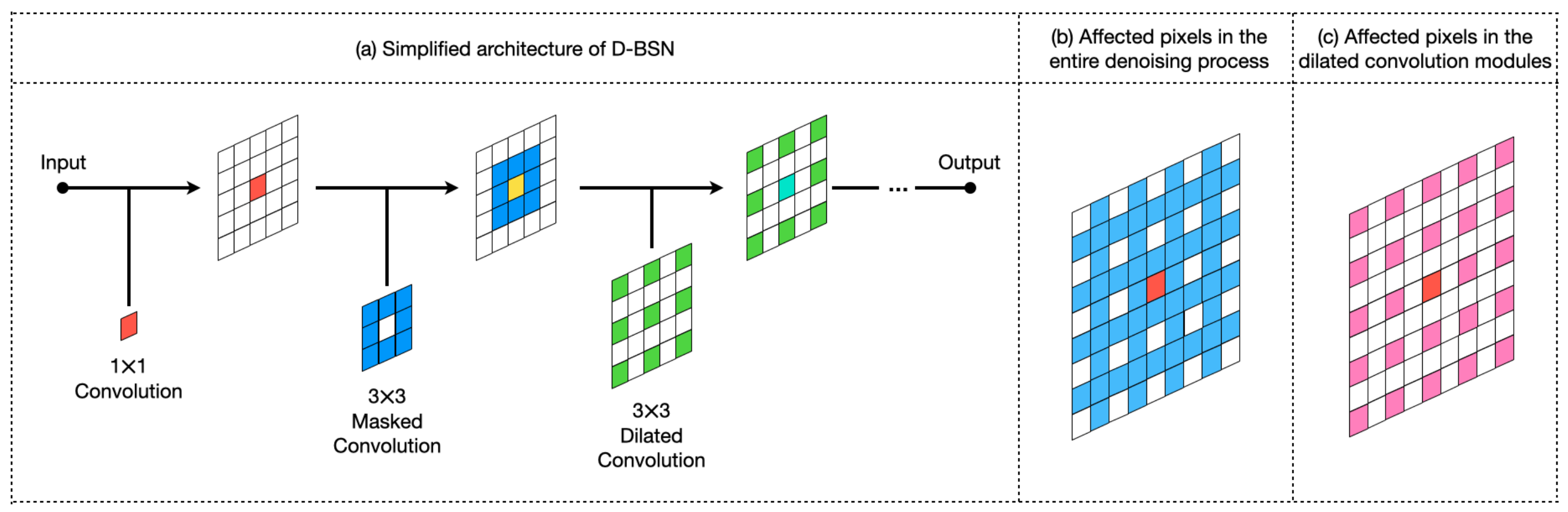}
    \\
    \vspace{-0mm}
    \caption{
        \textbf{The architecture of the dilated blind-spot network (D-BSN).} (a) Simplified architecture of D-BSN, (b) visualization of affected pixels (blue-colored) when the red-colored pixel is restored in the entire denoising process ($d=2$). (c) visualization of affected pixels (pink-colored) when the red-colored pixel is restored in the dilated convolution modules ($d=2$).
        }
    \label{fig:fig2}
    \vspace{-0mm}
\end{figure*}

\section{Background} 
\subsection{Revisiting the Dilated Blind-Spot Network}
This section introduces the blind-spot strategy~\cite{N2V} and D-BSN~\cite{DBSN}. The blind-spot strategy plays an essential role in numerous self-supervised denoising methods. The principle of the blind-spot strategy is to mask a pixel in the receptive field. Then, a neural network reconstructs the masked pixel using adjacent pixels' information. This mechanism is applied to all pixels in the image. A neural network with the blind-spot strategy is trained by solving the following: 
\begin{linenomath}
    \begin{align}
        argmin_\theta \Sigma_iL(f_\theta(x_i'),x_i) ,
    \end{align}
\end{linenomath}
where $x_i'$ and $x_i$ are the $i$th input images with and without blind spots, respectively, and $L(\cdot)$ denotes the loss function (e.g., L1 loss). In addition, $f_{\theta}(\cdot)$ denotes the blind-spot network parameterized by $\theta$. Note that the masked input pixel value should not directly or indirectly affect the reconstruction. If the input pixel value affects the reconstruction, a neural network is trained to mimic the input image. It is called identity mapping. To avoid identity mapping, in previous studies, input pixels are substituted with the adjacent pixels~\cite{N2V} or dropped out using Bernoulli sampling~\cite{S2S}. 

\begin{figure*}[t]
    \renewcommand{\wp}{\linewidth}
    \centering
    \includegraphics[width=1.0\linewidth]{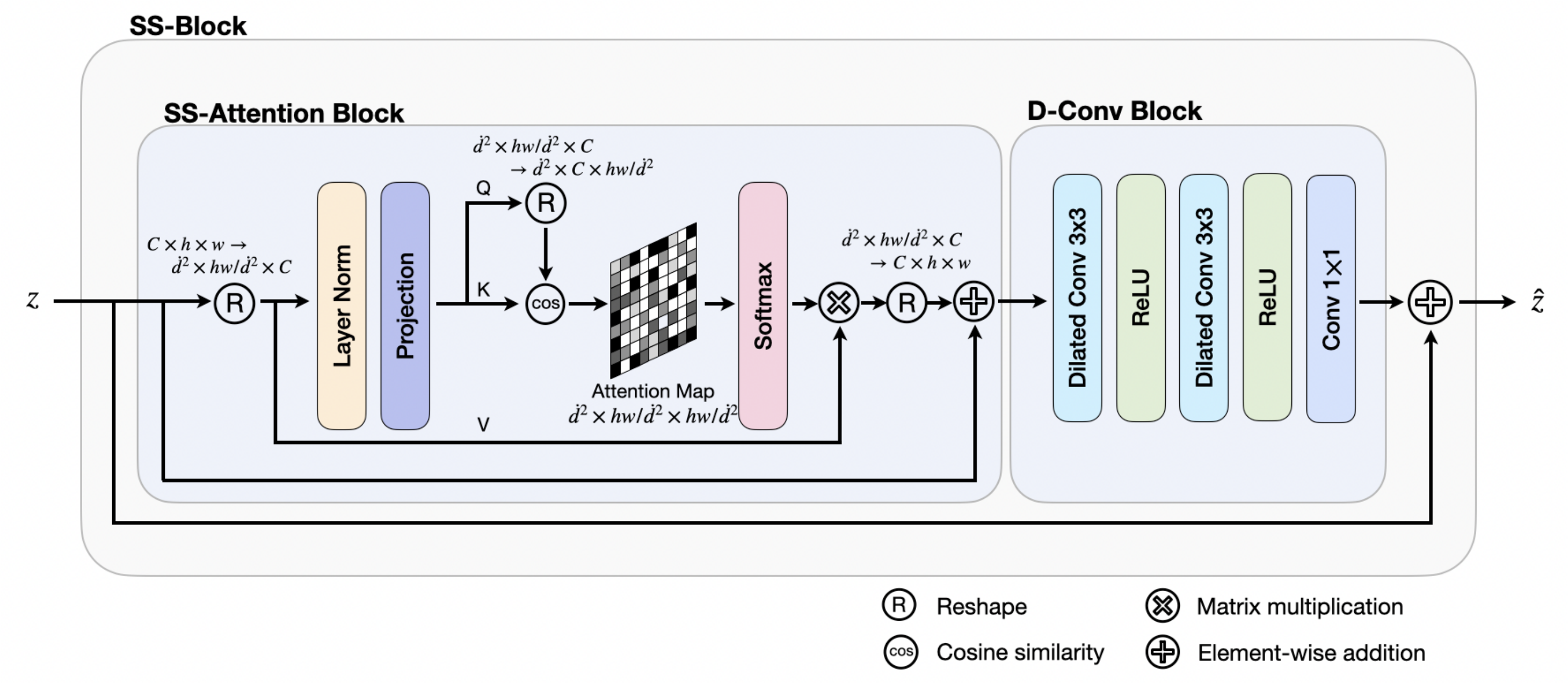}
    \\
    \vspace{0mm}
    \caption{
        \textbf{The architecture of our SS-Block which consists of SS-Attention and D-ConvBlock.} Our SS-Attention block is a lightweight self-attention module that can capture self-similarities, and D-ConvBlock, consisting of dilated convolution, activation layers, and $1\times1$ convolution layer, serves as a feed-forward network of the classic Transformer model. 
    }
    \label{fig:fig3}
    \vspace{-0mm}
\end{figure*}

The D-BSN proposed in~\cite{DBSN} is one of the self-supervised denoising methods with a blind-spot strategy. Specifically, D-BSN consists of three essential parts: $1\times1$ convolutional modules, a masked convolutional module, and dilated convolutional modules. Figure 2 depicts the D-BSN architecture. The $1\times1$ convolution modules perform feature extraction and aggregation per pixel. For the hidden embedding feature extracted by the $1\times1$ convolution, masked convolution generates blind spots by filtering after assigning zero to the center element of the convolutional filter. Specifically, applying masked convolution with a $k_{mc}$-sized kernel $w$ can be written as follows:
\begin{linenomath}
    \begin{gather}
        h^{(i+1)}=h^{(i)}*(w\otimes m)+b , \\
        m_{x,y} =
            \begin{cases}
                0, & \text{if } x=\lfloor k_{mc}/2 \rfloor \text{ and } y=\lfloor k_{mc}/2 \rfloor , \\
                1, & \text{otherwise},
            \end{cases}
    \end{gather}
\end{linenomath}
where $h^{(i)}$ is the hidden embedding of the $i$th layer, and $b$ denotes the bias of the layer. Additionally, $*$ and  $\otimes$ denote the convolutional operator and element-wise multiplication, respectively. After applying the masked convolution, noisy pixels are reconstructed using the information from adjacent pixels without using the masked pixel by applying stacked dilated convolutional modules. The dilation $d$ of the dilated convolution is determined as follows:
\begin{linenomath}
    \begin{align}
        d=(k_{mc}+1)/2 .
    \end{align}
\end{linenomath}
Because of these architectural characteristics of the D-BSN, even if a sufficient number of dilated convolutional modules are applied, each pixel is not affected by all pixels when reconstructed. In Figure 2, (b) and (c) present these architectural characteristics of the D-BSN when $k_{mc}=3$. In addition, (b) depicts pixels that affect when the masked pixel (red pixel) is restored in the entire process, and (c) illustrates pixels that affect when the masked pixel is restored in the dilated convolutional modules. Reconstructing the pixel $z(x,y)$ with successive $l$ dilated convolutional modules can be written as follows:
\begin{linenomath}
    \begin{align}
        z^{(i+l)}(x,y)=f_{\theta_d}(\{z^{(i)}(x+n_x d,y+n_y d) \nonumber \\
        ||n_x |\le l, |n_y| \le l\}),
    \end{align}
\end{linenomath}
where $n_x$ and $n_y$ denote the set of integers whose absolute values are not greater than $l$, $f_{\theta_d}(\cdot)$ represents dilated convolutional modules parameterized by $\theta _d$, and $z^{(i)}$ indicates the input of the $i$th dilated convolutional module. Thus, these architectural characteristics should be considered when designing a self-attention module in a pixel-wise manner using nonlocal information based on the D-BSN architecture. Without this consideration, the masked pixel directly or indirectly affects reconstruction for itself, and training the neural network may become unstable or fail due to identity mapping.
%

\subsection{Nonlocal Self-Similarity}

In general, natural images often have repetitive patterns. Using these repetitive patterns spread throughout an image is an effective image denoising method. We indicate this prior as nonlocal self-similarity. The denoising method using nonlocal self-similarity was first proposed in nonlocal means~\cite{NLM} and showed better performance over the conventional methods using local self-similarity. Afterward, various extensions~\cite{BM3D,peyre:hal-00419791,Xu_2015_ICCV,N2Sim} have been proposed using nonlocal self-similarity, and the methods have shown promising performance even though the studies used non-learning-based methods. Recently, CNN-based denoising methods have primarily been conducted. However, most CNN-based methods do not consider using nonlocal self-similarity because the notion of nonlocal self-similarity conflicts with the local filtering concept in the CNN.

\section{Method} 

Our main goal is to develop an effective D-BSN-based architecture that can perform the denoising task using a self-attention module that can consider nonlocal self-similarity. The challenge to achieving this goal is that the classic self-attention module has a high computational complexity to use in a pixel-wise manner. To alleviate the computational complexity, we designed a simplified self-attention module, focusing on obtaining nonlocal self-similarity. Furthermore, unlike the feedforward layers of a conventional vision transformer consisting of two fully connected layers, we adopt feedforward blocks consisting of two dilated convolutional layers and an $1\times1$ convolutional layer to reduce computational complexity. In this section, we first present our self-attention module, SS-Attention which is the core component of our proposed architecture. Subsequently, we present our D-BSN-based architecture which can be trained in a fully self-supervised manner.
\begin{figure}[!bt]
    \renewcommand{\wp}{\linewidth}
    \centering
    \includegraphics[width=1.0\linewidth]{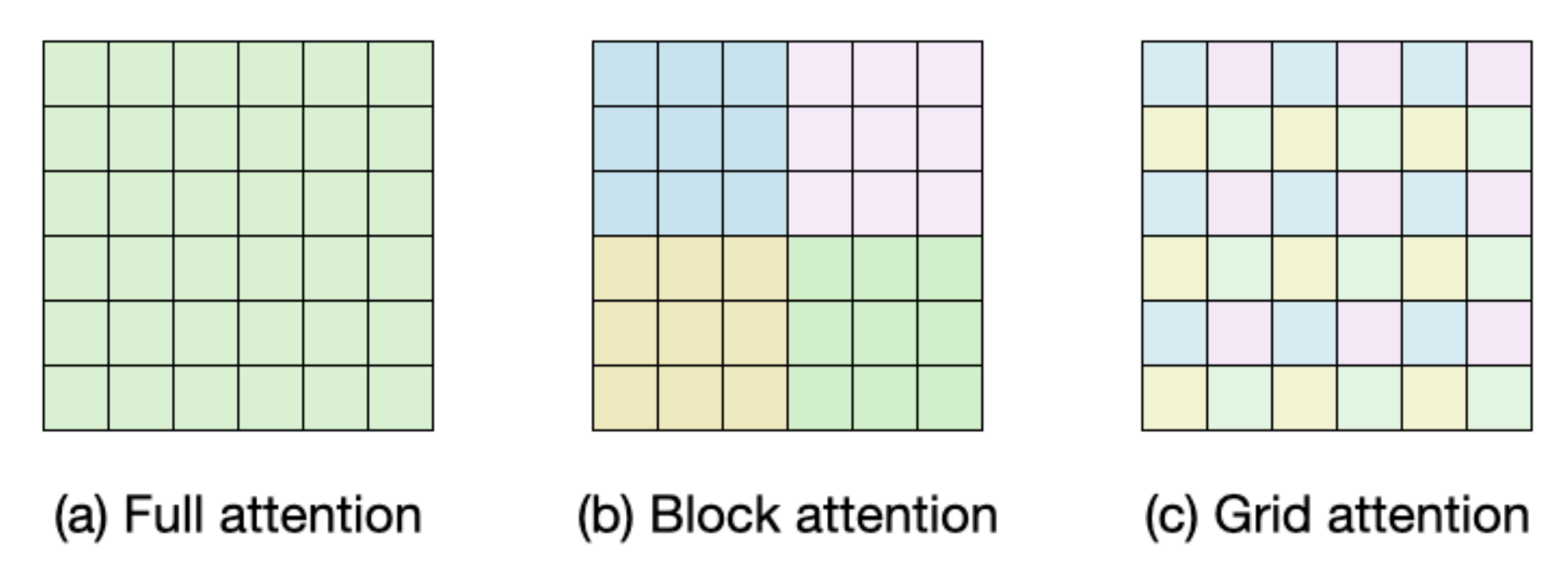}
    \\
    \vspace{0mm}
    \caption{
        \textbf{An illustration of self-attention schemes.} The same colored pixels are mixed by the self-attention modules.
    }
    \label{fig:fig4}
    \vspace{-0mm}
\end{figure}

\begin{figure*}[!htb]
    \renewcommand{\wp}{\linewidth}
    \centering
    \includegraphics[width=0.90\linewidth]{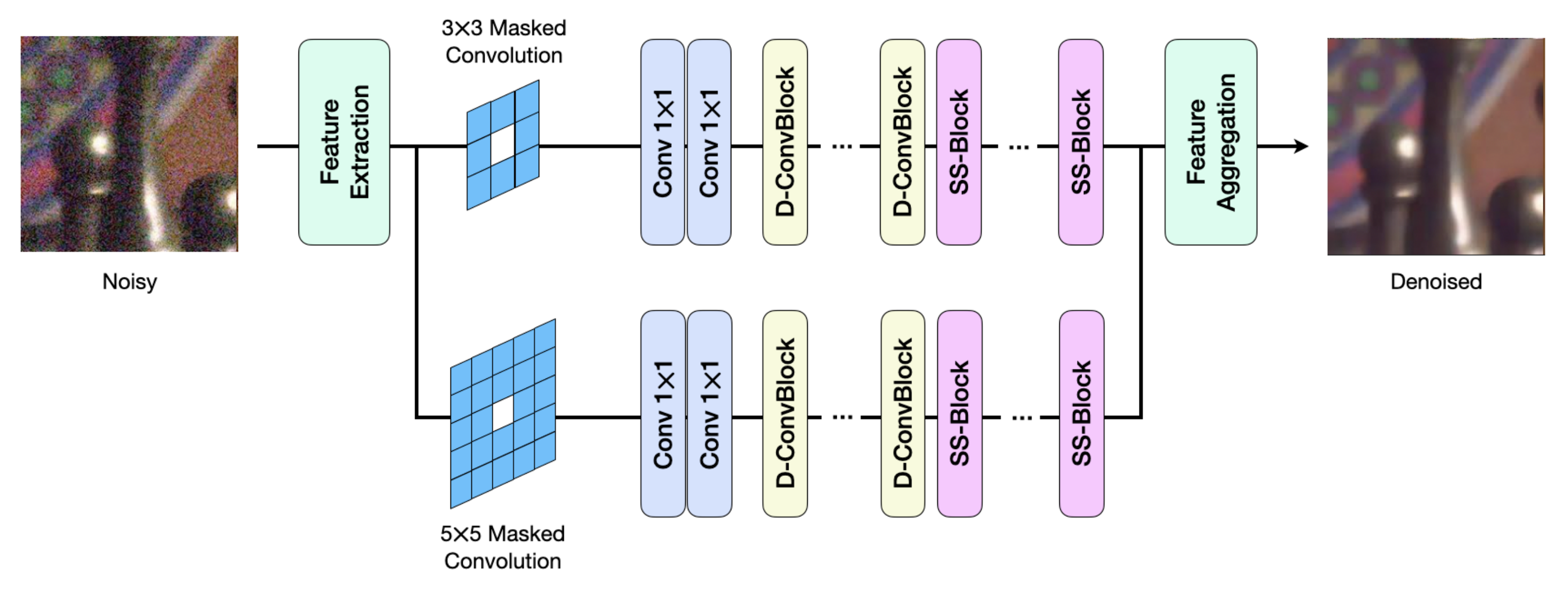}
    \\
     \vspace{-0mm}
    \caption{
        \textbf{The overall architecture of SS-BSN.} There are two paths for dilated convolution modules starting with $3\times3$ and $5\times5$ masked convolutions, respectively. For each path, we stack a total of 9 dilated convolution modules, the first $(9-m)$ modules are DConvBlocks, and the following $m$ modules are SS-Blocks. The feature extraction and feature aggregation modules consist of 1$\times1$ convolution layers and activation layers (ReLU). In our experiments, $m$ is set to 3.
    }
    \label{fig:fig5}
     \vspace{-0mm}
\end{figure*}

\subsection{Self-Similarity-Based Attention}

The architecture of SS-Attention is presented in Figure 3. The SS-Attention module first generates the self-similarity-based attention map, applies the attention mechanism, and propagates the embeddings through the dilated convolution block (D-ConvBlock). This section introduces the details of SS-Attention architecture and the considerations when designing this module. The classic self-attention module has a limitation when performed in a pixel-wise manner due to its complexity, which quadratically increases with spatial resolution. We simplify the self-attention module to reduce the computational complexity. The computational complexity of the classic self-attention mechanism, the so-called multi-head self-attention (MSA), is provided below:
\begin{linenomath}
    \begin{align}
        \mathcal{O}(MSA)=4hwC+2(hw)^2C,
    \end{align}
\end{linenomath}
where $h$ and $w$ indicate the dimensions of the spatial resolution, and $C$ denotes the channels of the tensor. The left term represents the complexity of applying four linear transforms, and the right term indicates the complexity of generating and applying an attention map. As the equation reveals, the major computational overhead of the MSA is from the size of the attention map. Therefore, reducing the size of the attention map is key to reducing the computational complexity of the self-attention module. 
As mentioned in Section 2, in D-BSN-based architecture, there are sets of pixels that affect each other during the reconstruction process. Therefore, a self-attention map should be generated within a pixel set that can affect each other, defined by Eq. (5). To achieve this, we adopted the grid attention~\cite{MaxViT}. We first reshape the tensor of shape ($C\times h \times w$) into shape ($d^2\times hw/d^2\times C$) using a $d\times d$ grid. Then, we employ self-attention on the decomposed tensor. Through this process, we generate the attention map $A\in\mathbb{R}^{hw/d^2\times hw/d^2}$ for each set of pixels. Figure 4 compares the full attention, block attention~\cite{SwinIR,IPT}, and grid attention.
The attention map we generated is much smaller than the attention map for MSA. However, this reduction may be insufficient in environments with limited hardware. Therefore, we used the parameter $\gamma$ to resize the grid. In general, in the case of natural images, repeated patterns appear globally, so even if more sparsity is added to the self-attention modules, the performance of the denoising module does not degrade much. Thus, the final grid size $\dot{d}$ is determined by $\dot{d}=\gamma \cdot d$, and $\dot{d}^2$ number of $A$ tensors are generated. To summarize, reconstructing the pixel $z(x,y)$ with an SS-Attention block can be written as follows:
\begin{linenomath}
    \begin{align}
        \hat{z}(x,y)=f_{\theta_{ss}}(\{z(x+n_x \dot{d},y+n_y \dot{d})\}),
    \end{align}
\end{linenomath}
where $n_x$ and $n_y$ denote a set of integers, and $f_{\theta_{ss}}(\cdot )$ denotes a SS-Attention block. 
In addition to generating and applying the attention map, the MSA consists of four linear transforms for generating a query ($Q$), key ($K$), value ($V$), and output. To further simplify the self-attention module, we design a self-attention mechanism using a linear transform to reduce the linear transform-related complexity $\mathcal{O}(4hwC)$ to $\mathcal{O}(hwC)$. Specifically, we integrate $Q$ with $K$, and the gridded input tensor serves as $V$. This design makes SS-Attention focus on capturing nonlocal self-similarities and improves training stability. From a normalized tensor $Y\in \mathbb{R}^{\dot{d}^2\times hw/\dot{d}^2\times C}$ and gridded input tensor $\hat{z}\in \mathbb{R}^{\dot{d}^2\times hw\dot{d}^2 \times C}$, SS-Attention generates $Q$, $K$, and $V$ as follows:
\begin{linenomath}
\begin{align}
Q=YW_{qk}, K=YW_{qk}, V=\hat{z} \nonumber , \\
Q,K,V\in\mathbb{R}^{\dot{d}^2\times hw/\dot{d}^2\times C},
\end{align}
\end{linenomath}
where $W_{qk}\in\mathbb{R}^{C\times C}$ denotes a linear matrix. With $Q$, $K$, and $V$ generated in this way, the process of SS-Attention is defined as follows:
\begin{linenomath}
\begin{align}
z^{(l+1)}=Softmax(\frac{1+cos(Q,K^{T})}{\sqrt{C}})V + z^{(l)},
\end{align}
\end{linenomath}
where $z^{(l)}$ denotes the input of the $l$th SS-Attention block. Overall, the computational complexity of SS-Attention is provided below:
\begin{linenomath}
\begin{align}
\mathcal{O}(\text{SS-Attention})=hwC + \frac{2(hw)^2C}{\dot{d}^2} .
\end{align}
\end{linenomath}
In our experimental settings ($\gamma=2$), the average computational complexity of our SS-Attention is only about $3.8\%$ of MSA.
In addition, previous studies related to nonlocal self-similarity compared similarities between patches. However, in this study, pixel-wise features are compared because the information for the adjacent pixels is embedded in the central pixel due to the convolutional operations.

\newcommand{\ours}{^\dagger}
\newcommand{\ensemble}{_e}

\begin{table*}[!t]
  \small
  \renewcommand\arraystretch{1.1}
  \begin{center}
    \label{table:raw}
    \vspace{0mm}
    \scalebox{1.0}{
    \begin{tabular}{cccccc}
      \toprule
      & \multirow{2}{*}{\begin{tabular}[c]{@{}c@{}}Method \end{tabular}} 
      & SIDD~\cite{SIDD} & DND~\cite{DND} \\
      && PSNR$^\uparrow$(dB) / SSIM$^\uparrow$ & PSNR$^\uparrow$ (dB) / SSIM$^\uparrow$   \\

      \midrule
      \multirow{2}{*}{\begin{tabular}[c]{@{}c@{}}Non-learning Based \end{tabular}}
      & BM3D~\cite{BM3D}  & 25.65 / 0.685 & 34.51 / 0.851 \\
      & WNNM~\cite{WNNM}  & 25.78 / 0.809 & 34.67 / 0.865 \\

      \midrule
      \multirow{2}{*}{\begin{tabular}[c]{@{}c@{}}Supervised \end{tabular}}
      & DnCNN ~\cite{DnCNN}  & 36.63 / 0.920$\ours$ & 38.00 / 0.934$\ours$ \\
      & DANet~\cite{DANet}  & 39.46 / 0.956 & 39.47 / 0.955 \\

      \midrule
      \multirow{2}{*}{\begin{tabular}[c]{@{}c@{}}Supervised\\(Synthetic pairs)\ \end{tabular}}
      & CBDNet~\cite{CBDNet}  & 33.28 / 0.868 & 38.05 / 0.942 \\
      & Zhou et al.~\cite{Zhou}  & 34.02 / 0.898$\ours$ & 38.40 / 0.945 \\

      \midrule
      \multirow{5}{*}{\begin{tabular}[c]{@{}c@{}}Self-Supervised \end{tabular}}
      & Noise2Void~\cite{N2V}  & 27.68 / 0.668 & - \\
      & Noise2Self~\cite{N2S}  & 29.59 / 0.808 & - \\
      & R2R~\cite{R2R}  & 34.78 / 0.898 & - \\
      & AP-BSN ~\cite{APBSN}  & 35.97 / 0.909$\ours$ & 37.46 / 0.924 \\
      & AP-BSN$\ensemble$ ~\cite{APBSN}  & 37.05 / 0.934 & 38.09 / 0.937 \\

      \midrule
      \multirow{2}{*}{\begin{tabular}[c]{@{}c@{}}Ours\\(Self-Supervised) \end{tabular}}
      & \textbf{SS-BSN}  & \textbf{36.73 / 0.923} & \textbf{37.72 / 0.928} \\
      & \textbf{SS-BSN$\ensemble$}  & \textbf{37.42 / 0.937} & \textbf{38.46 / 0.940} \\
      \bottomrule
    \end{tabular}}
    \caption{\textbf{Quantitative results on SIDD and DND datasets.} By default, the baseline results of benchmark datasets are cited from the official website for a fair comparison. We report our experimental results when the results are not reported on the benchmark websites. $\dagger$ indicates our experimental result, and $e$ denotes the methods which adopt the self-ensemble strategy proposed in AP-BSN.}
  \end{center}
  \vspace{0mm}
\end{table*}

\subsection{Self-Similarity-Based Blind-Spot Network}
We integrate SS-Attention into the blind-spot network (SS-BSN), which can be trained in a self-supervised manner. The proposed SS-BSN is inspired by the D-BSN~\cite{DBSN} and AP-BSN~\cite{APBSN}. Figure 5 illustrates the overall architecture of the SS-BSN. As mentioned in Section 2, to train a fully self-supervised denoising neural network, we first extract features of the image with $1\times1$ convolutions and generate blind pixels through masked convolutions. Subsequently, each pixel is reconstructed using information from the adjacent pixels through a D-ConvBlock and SS-Block. The D-ConvBlock consists of the remaining parts except for SS-Attention in the SS-Block in Figure 3. Finally, we take $1\times1$ convolutions for the tensors from the SS-Block to reduce the channels and generate the denoised image.
As depicted in Figure 5, we introduce successive D-ConvBlocks and SS-Blocks in the last $m$ layers. The SS-Attention, the main component of the SS-Block, determines self-similarity based on the cosine similarity of embedded features of each pixel; thus, it is inefficient to determine self-similarities by comparing noisy features that are not sufficiently denoised~\cite{Xu_2015_ICCV}. Therefore, we designed SS-BSN such that embedded features, which are sufficiently denoised through successive D-ConvBlocks, serve as input to the SS-Blocks for capturing nonlocal self-similarities. To justify this approach, we provide additional experimental results in Section 4.5.

\section{Experimental Results} 
\subsection{Experimental Settings} 
\paragraph{Dataset}
To evaluate the proposed method, we use the Smartphone Image Denoising Dataset (SIDD)~\cite{SIDD} and Darmstadt Noise Dataset (DND)~\cite{DND}. The SIDD medium split contains 320 noisy-clean image pairs taken in various lighting conditions and ISO using five different smartphones. We also adopted the SIDD validation and benchmark dataset for validation and testing. The SIDD validation and benchmark dataset contain 1,280 patches of size $256\times256$, each. 
The DND dataset contains 50 noisy images, each of which contains 20 bounding boxes of size $512\times512$. Four cameras capture noisy images under a higher ISO with a shorter exposure time. The DND dataset does not provide training and validation images; therefore, we used the DND dataset for training and performance evaluation. This experimental setting is possible because the SS-BSN can be trained in a fully self-supervised manner. 
Although the SIDD and DND datasets provide both sRGB and RAW data, we evaluate the denoising performance on the sRGB data. The ground truth images of the SIDD benchmark and DND dataset are not provided, but the peak signal-to-noise ratio (PSNR) and structural similarity index measure (SSIM) results for the denoising results can be obtained through the online submission system on the SIDD benchmark website\footnote{\url{https://www.eecs.yorku.ca/~kamel/sidd/benchmark.php}} and the DND benchmark website\footnote{\url{https://noise.visinf.tu-darmstadt.de/benchmark/}}.

\begin{figure*}[!tb]
    \renewcommand{\wp}{\linewidth}
    \centering
    \includegraphics[width=0.9\linewidth]{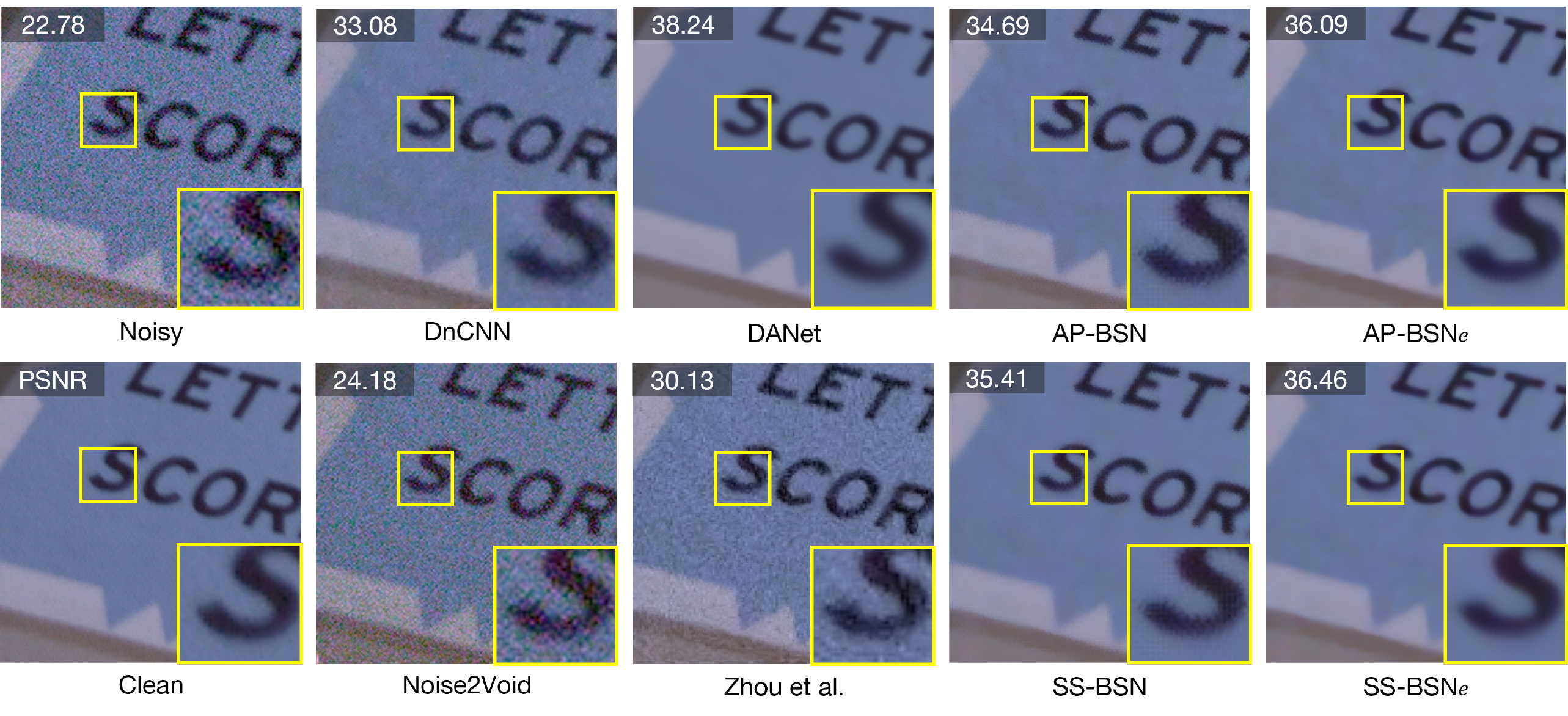}
    \\
     \vspace{-0mm}
    \caption{
        \textbf{Visual comparison of denoising sRGB images in the SIDD validation dataset.}
    }
    \label{fig:fig6}
     \vspace{-0mm}
\end{figure*}

\paragraph{Pixel-Shuffle Downsampling}
Naive D-BSN has pixel-wise independent noise assumption; thus, it is ineffective in removing real-world noise with spatial correlation. Recently, methods~\cite{Zhou,APBSN} that attempt to remove the spatial correlation of real-world noise using pixel-shuffle downsampling have been proposed. Fortunately, the D-BSN can effectively remove real-world noise by training with pixel-shuffle downsampled images. In particular, the method for minimizing the aliasing artifacts that can arise when applying pixel-shuffle downsampling using an asymmetric pixel-shuffle stride factor in the training and testing phases is proposed in the AP-BSN. This method shows promising performance. Thus, we adopt this approach to the proposed method and perform real-world denoising with a pixel-shuffle stride factor of 5 in the training phase and 2 in the testing phase.

\paragraph{Implementation Details} 
To optimize the SS-BSN\footnote{Our code is available at: \url{https://github.com/YoungJooHan/SS-BSN}}, we randomly extract the patches of size $120\times120$ from noisy images and augment all training images by randomly flipping and rotating them by $90^\circ$. In addition, we used the L1 loss and the Adam~\cite{ADAM} optimizer with an initial learning rate of $10^{-4}$. At the $16th$ epoch, the learning rate is multiplied by 0.1, where our model is trained over 20 epochs. We set $\gamma$ to 2 for the SS-Attention module and $m$ to 3 for the SS-BSN architecture. These hyperparameters are determined by our additional experiments described in Section 4.5.
%

\subsection{Results on Real-world Denoising}
Table 1 lists the PSNR/SSIM results of SS-BSN and various baselines. We follow the submission guidelines for the SIDD and DND datasets to evaluate the proposed method. By default, the baseline results on benchmark datasets are cited from official websites for a fair comparison. However, if the results are not reported on the benchmark websites, our experimental results are reported. 
The proposed method is compared to traditional non-learning- and learning-based methods in the experiments. Specifically, the methods we include for the comparison cover non-learning based methods (BM3D~\cite{BM3D} and WNNM~\cite{WNNM}), supervised denoising methods (DnCNN~\cite{DnCNN} and DANet~\cite{DANet}), supervised methods trained with generated synthetic noise (CBDNet~\cite{CBDNet} and Zhou et al.~\cite{Zhou}), and self-supervised methods (Noise2Void~\cite{N2V}, Noise2Sself~\cite{N2S}, R2R~\cite{R2R}, and AP-BSN~\cite{APBSN}). We also provide a qualitative comparison between SS-BSN and various baselines in Figure 6.
In Table 1, SS-BSN outperforms AP-BSN on the SIDD and DND datasets, which previously performed best in a self-supervised manner. Specifically, with a self-ensemble method, which is proposed in AP-BSN, SS-BSN obtains PSNR gains of 0.37 dB on both datasets; without a self-ensemble method, SS-BSN obtains PSNR gains of 0.76 dB and 0.26 dB over the AP-BSN method. Further, SS-BSN with the self-ensemble method obtains better PSNR values than the supervised methods using synthetic pairs, which have the constraint that a sufficient amount of clean images must be accessible.

\begin{table}[!bt]
\centering
\renewcommand{\arraystretch}{1.2}
\scalebox{1.0}{
\begin{tabular}{c | c | c | c | c}
\bf $\emph{SS}$ & \bf $\emph{QK}$ & \bf $\emph{CS}$ &  \bf $\emph{DF}$ & \bf \textsc{PSNR/SSIM} \\
\hline
$\times$ & $\times$ & $\times$ & $\times$ 	& 35.97/0.837 \\
$\surd$ & $\surd$ & $\times$ & $\times$ 	& 35.96/0.837 \\
$\surd$ & $\times$ & $\surd$ & $\times$ 	& 36.04/0.839 \\
$\surd$ & $\surd$ & $\surd$ & $\times$ 	& 36.26/0.850 \\
$\surd$ & $\times$ & $\surd$ & $\surd$ 	& 36.68/0.857 \\
$\surd$ & $\surd$ & $\surd$ & $\surd$ 	& {\bf 36.78/0.860} \\
\end{tabular}
}
\caption{
\textbf{Ablation study of SS-Attention and SS-BSN on SIDD validation dataset.} \textbf{SS} denotes the SS-Attention, \textbf{QK} the query key integration, \textbf{CS} the cosine similarity, and \textbf{DF} the determination of self-similarities on the denoised features. Specifically, in the experiment labeled with DF, the last three D-ConvBlocks are replaced with SS-Blocks. 
}
\end{table}
\begin{figure*}[!tbh]
    \renewcommand{\wp}{\linewidth}
    \centering
    \includegraphics[width=1.0\linewidth]{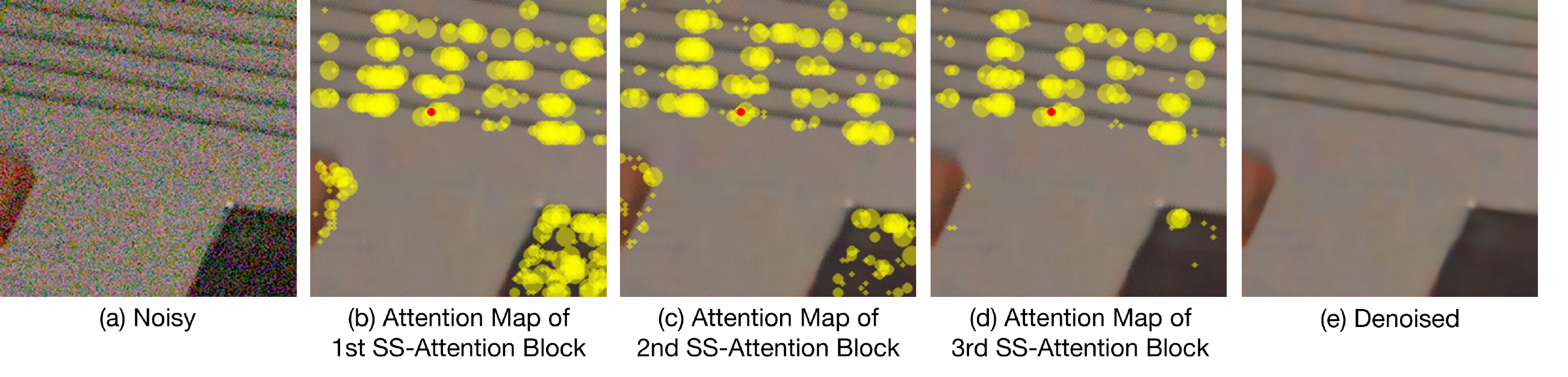}
     \vspace{-0mm}
     \hspace{-2mm}
    \caption{
        \textbf{Visualization of the self-similarity-based-attention maps.} (a) Noisy input image, (b)-(d) visualization of the self-similarity-based-attention maps from different SS-Attention blocks. (e) denoised estimates of our method. In (b)-(d), the yellow circles show the pixels with high similarity with the masked pixel (red circle) predicted by our method. A larger radius of the yellow circles indicates high similarity.
    }
    \label{fig:fig7}
     \vspace{-0mm}
\end{figure*}

\subsection{Ablation Study} 
{
Table 2 summarizes the performance of different architecture choices for our proposed SS-Attention and SS-BSN. In the experiment that applied only query key integration or cosine similarity to SS-Attention, no significant performance improvement is observed compared to the baseline. However, in the experiment where both query key integration and cosine similarity are applied, a meaningful performance improvement is observed. We also find that determining self-similarities on denoised features significantly improves the denoising performance.
}

\subsection{Visualization of SS-Attention Results} 
The visualization results of the self-similarity-based attention maps from different SS-Attention blocks are shown in Figures 1 and 7. To justify the effectiveness of the SS-Attention mechanism, we used a visualization method similar to that in~\cite{DCN}. As shown in Figure 7, the self-similarity-based attention map of the first SS-Block (Figure 7b) is very noisy and does not represent meaningful information since the attention map is generated by using the output of the D-ConvBlock. However, in the deep block (Figure 7d), the self-similarity-based attention maps from SS-Attention blocks represent accurate and meaningful information. Yellow circles are drawn on pixels with high self-similarity attention values, and a larger radius of yellow circles indicates high similarity. 

\begin{figure}[!tb]
    \renewcommand{\wp}{\linewidth}
    \centering
    \includegraphics[width=1.0\linewidth]{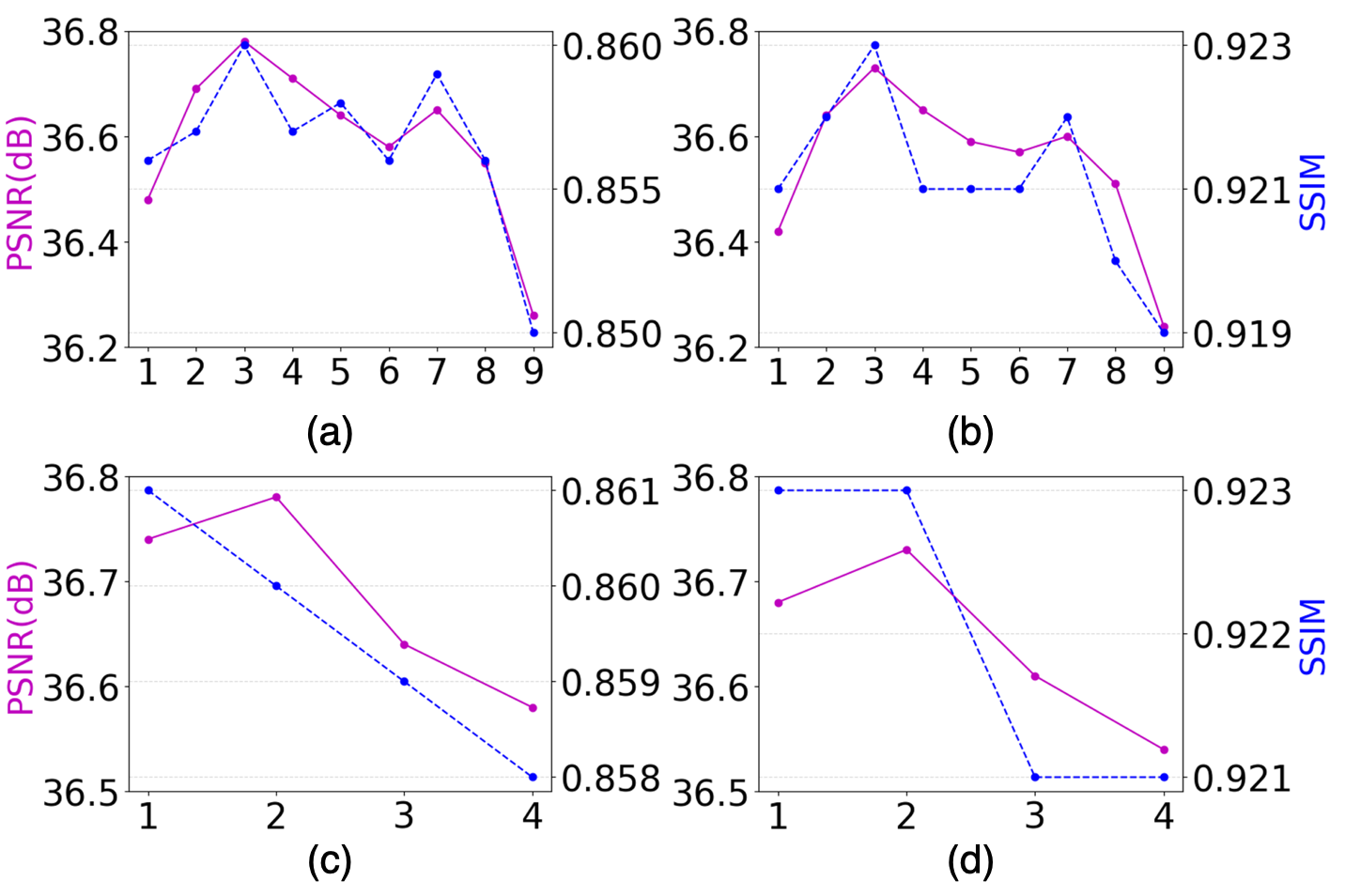}
     \vspace{-0mm}
     \hspace{-2mm}
    \caption{
        \textbf{The analysis of hyperparameters.} (a), (b) Performance comparison according to hyperparameter $m$ in SIDD validation and benchmark dataset, respectively. (c), (d) Performance comparison according to hyperparameter $\gamma$ in SIDD validation and benchmark dataset, respectively. 
    }
    \label{fig:fig7}
     \vspace{-0mm}
\end{figure}
\subsection{Hyperparameters} 
The SS-BSN and SS-Attention have hyperparameters of $\gamma$ and $m$, respectively. Figure 8 shows the effect of these hyperparameters on the performance on the SIDD validation and benchmark datasets. The hyperparameter of SS-BSN, $m$ determines the number of the last $m$ dilated convolution modules that SS-Blocks will be substituted. Since our proposed SS-Attention is based on the similarity between pixel features, it may be ineffective to calculate the cosine similarity between noisy pixels. Therefore, features that are denoised from D-ConvBlocks are used as input to the SS-Block. Figure 8a and 8b show the performance comparison when the last $m$ blocks are substituted with SS-Blocks. The experimental results show that it is effective when SS-Block is applied to denoised features. To this end, we set $m$ to 3 in our experiments.
The hyperparameter of SS-Attention, $\gamma$ determines the size of the attention map, which greatly affects the computational complexity of SS-Block.  A large attention map of self-attention module in a pixel-wise manner means that numerous pixels are considered during the reconstruction. However, in general, since repetitive patterns appear globally, adding sparsity to the attention map does not significantly degrade the denoising performance. It may be seen from Figure 8c and 8d that performance degradation is not noticeable until $\gamma$ is 2, but the computational complexity drops dramatically. Therefore, we set $\gamma$ to 2 in our experiments.

\section{Conclusion} 

This paper presents SS-Attention, a novel self-similarity-based self-attention module that can capture long-range dependency and obtain information from nonlocal self-similarities. Furthermore, we integrate SS-Attention into the blind-spot network (SS-BSN), which can be trained in a fully self-supervised manner. This paper focused on designing a lightweight self-attention module that can be trained in a pixel-wise manner. The experiments demonstrate the effectiveness of the proposed model over various baselines in real-world denoising. Additionally, we provide justification for our SS-Attention with the visualization of self-similarity-based attention maps. In the future, we hope our work can be a key to solving the challenging points of self-supervised denoising.
%

\section*{Acknowledgments}


%
This work was supported by the National Research Foundation of Korea(NRF) grant funded by the Korea government. (MSIT) (2023R1A2C1005744)
%

\bibliographystyle{named}
\bibliography{SSBSN}

\begin{thebibliography}{}

\bibitem[\protect\citeauthoryear{Abdelhamed \bgroup \em et al.\egroup
  }{2018}]{SIDD}
Abdelrahman Abdelhamed, Stephen Lin, and Michael~S. Brown.
\newblock A high-quality denoising dataset for smartphone cameras.
\newblock In {\em Proceedings of the IEEE Conference on Computer Vision and
  Pattern Recognition (CVPR)}, June 2018.

\bibitem[\protect\citeauthoryear{Batson and Royer}{2019}]{N2S}
Joshua Batson and Loic Royer.
\newblock {N}oise2{S}elf: Blind denoising by self-supervision.
\newblock In Kamalika Chaudhuri and Ruslan Salakhutdinov, editors, {\em
  Proceedings of the 36th International Conference on Machine Learning},
  volume~97 of {\em Proceedings of Machine Learning Research}, pages 524--533.
  PMLR, 09--15 Jun 2019.

\bibitem[\protect\citeauthoryear{Buades \bgroup \em et al.\egroup }{2005}]{NLM}
A.~Buades, B.~Coll, and J.-M. Morel.
\newblock A non-local algorithm for image denoising.
\newblock In {\em 2005 IEEE Computer Society Conference on Computer Vision and
  Pattern Recognition (CVPR'05)}, volume~2, pages 60--65 vol. 2, 2005.

\bibitem[\protect\citeauthoryear{Chen \bgroup \em et al.\egroup }{2021}]{IPT}
Hanting Chen, Yunhe Wang, Tianyu Guo, Chang Xu, Yiping Deng, Zhenhua Liu, Siwei
  Ma, Chunjing Xu, Chao Xu, and Wen Gao.
\newblock Pre-trained image processing transformer.
\newblock In {\em Proceedings of the IEEE/CVF Conference on Computer Vision and
  Pattern Recognition (CVPR)}, pages 12299--12310, June 2021.

\bibitem[\protect\citeauthoryear{Dabov \bgroup \em et al.\egroup }{2007}]{BM3D}
Kostadin Dabov, Alessandro Foi, Vladimir Katkovnik, and Karen Egiazarian.
\newblock Image denoising by sparse 3-d transform-domain collaborative
  filtering.
\newblock {\em IEEE Transactions on Image Processing}, 16(8):2080--2095, 2007.

\bibitem[\protect\citeauthoryear{Dai \bgroup \em et al.\egroup }{2017}]{DCN}
Jifeng Dai, Haozhi Qi, Yuwen Xiong, Yi~Li, Guodong Zhang, Han Hu, and Yichen
  Wei.
\newblock Deformable convolutional networks.
\newblock In {\em Proceedings of the IEEE International Conference on Computer
  Vision (ICCV)}, Oct 2017.

\bibitem[\protect\citeauthoryear{Dosovitskiy \bgroup \em et al.\egroup
  }{2020}]{ViT}
Alexey Dosovitskiy, Lucas Beyer, Alexander Kolesnikov, Dirk Weissenborn,
  Xiaohua Zhai, Thomas Unterthiner, Mostafa Dehghani, Matthias Minderer, Georg
  Heigold, Sylvain Gelly, Jakob Uszkoreit, and Neil Houlsby.
\newblock An image is worth 16x16 words: Transformers for image recognition at
  scale, 2020.

\bibitem[\protect\citeauthoryear{Gu \bgroup \em et al.\egroup }{2014}]{WNNM}
Shuhang Gu, Lei Zhang, Wangmeng Zuo, and Xiangchu Feng.
\newblock Weighted nuclear norm minimization with application to image
  denoising.
\newblock In {\em 2014 IEEE Conference on Computer Vision and Pattern
  Recognition}, pages 2862--2869, 2014.

\bibitem[\protect\citeauthoryear{Guo \bgroup \em et al.\egroup }{2019}]{CBDNet}
Shi Guo, Zifei Yan, Kai Zhang, Wangmeng Zuo, and Lei Zhang.
\newblock Toward convolutional blind denoising of real photographs.
\newblock In {\em Proceedings of the IEEE/CVF Conference on Computer Vision and
  Pattern Recognition (CVPR)}, June 2019.

\bibitem[\protect\citeauthoryear{Kingma and Ba}{2015}]{ADAM}
Diederik~P Kingma and Jimmy Ba.
\newblock Adam: A method for stochastic optimization.
\newblock In {\em ICLR}, 2015.

\bibitem[\protect\citeauthoryear{Krull \bgroup \em et al.\egroup }{2019}]{N2V}
Alexander Krull, Tim-Oliver Buchholz, and Florian Jug.
\newblock Noise2void - learning denoising from single noisy images.
\newblock In {\em 2019 IEEE/CVF Conference on Computer Vision and Pattern
  Recognition (CVPR)}, pages 2124--2132, 2019.

\bibitem[\protect\citeauthoryear{Lee \bgroup \em et al.\egroup }{2022}]{APBSN}
Wooseok Lee, Sanghyun Son, and Kyoung~Mu Lee.
\newblock Ap-bsn: Self-supervised denoising for real-world images via
  asymmetric pd and blind-spot network.
\newblock In {\em Proceedings of the IEEE/CVF Conference on Computer Vision and
  Pattern Recognition (CVPR)}, pages 17725--17734, June 2022.

\bibitem[\protect\citeauthoryear{Lehtinen \bgroup \em et al.\egroup
  }{2018}]{N2N}
Jaakko Lehtinen, Jacob Munkberg, Jon Hasselgren, Samuli Laine, Tero Karras,
  Miika Aittala, and Timo Aila.
\newblock {N}oise2{N}oise: Learning image restoration without clean data.
\newblock In Jennifer Dy and Andreas Krause, editors, {\em Proceedings of the
  35th International Conference on Machine Learning}, volume~80 of {\em
  Proceedings of Machine Learning Research}, pages 2965--2974. PMLR, 10--15 Jul
  2018.

\bibitem[\protect\citeauthoryear{Liang \bgroup \em et al.\egroup
  }{2021}]{SwinIR}
Jingyun Liang, Jiezhang Cao, Guolei Sun, Kai Zhang, Luc Van~Gool, and Radu
  Timofte.
\newblock Swinir: Image restoration using swin transformer.
\newblock In {\em Proceedings of the IEEE/CVF International Conference on
  Computer Vision (ICCV) Workshops}, pages 1833--1844, October 2021.

\bibitem[\protect\citeauthoryear{{Niu} \bgroup \em et al.\egroup
  }{2020}]{N2Sim}
Chuang {Niu}, Mengzhou {Li}, Fenglei {Fan}, Weiwen {Wu}, Xiaodong {Guo}, Qing
  {Lyu}, and Ge~{Wang}.
\newblock {Suppression of Correlated Noise with Similarity-based Unsupervised
  Deep Learning}.
\newblock {\em arXiv e-prints}, page arXiv:2011.03384, November 2020.

\bibitem[\protect\citeauthoryear{Pang \bgroup \em et al.\egroup }{2021}]{R2R}
Tongyao Pang, Huan Zheng, Yuhui Quan, and Hui Ji.
\newblock Recorrupted-to-recorrupted: Unsupervised deep learning for image
  denoising.
\newblock In {\em Proceedings of the IEEE/CVF Conference on Computer Vision and
  Pattern Recognition (CVPR)}, pages 2043--2052, June 2021.

\bibitem[\protect\citeauthoryear{Peyr{\'e} \bgroup \em et al.\egroup
  }{2011}]{peyre:hal-00419791}
Gabriel Peyr{\'e}, S{\'e}bastien Bougleux, and Laurent~D. Cohen.
\newblock {Non-local Regularization of Inverse Problems}.
\newblock {\em {Inverse Problems and Imaging }}, 5(2):511--530, 2011.

\bibitem[\protect\citeauthoryear{Plotz and Roth}{2017}]{DND}
Tobias Plotz and Stefan Roth.
\newblock Benchmarking denoising algorithms with real photographs.
\newblock In {\em Proceedings of the IEEE Conference on Computer Vision and
  Pattern Recognition (CVPR)}, July 2017.

\bibitem[\protect\citeauthoryear{Quan \bgroup \em et al.\egroup }{2020}]{S2S}
Yuhui Quan, Mingqin Chen, Tongyao Pang, and Hui Ji.
\newblock Self2self with dropout: Learning self-supervised denoising from
  single image.
\newblock In {\em Proceedings of the IEEE/CVF Conference on Computer Vision and
  Pattern Recognition (CVPR)}, June 2020.

\bibitem[\protect\citeauthoryear{Tu \bgroup \em et al.\egroup }{2022}]{MaxViT}
Zhengzhong Tu, Hossein Talebi, Han Zhang, Feng Yang, Peyman Milanfar, Alan
  Bovik, and Yinxiao Li.
\newblock Maxvit: Multi-axis vision transformer.
\newblock In Shai Avidan, Gabriel Brostow, Moustapha Ciss{\'e}, Giovanni~Maria
  Farinella, and Tal Hassner, editors, {\em Computer Vision -- ECCV 2022},
  pages 459--479, Cham, 2022. Springer Nature Switzerland.

\bibitem[\protect\citeauthoryear{Vaswani \bgroup \em et al.\egroup
  }{2017}]{Transformer}
Ashish Vaswani, Noam Shazeer, Niki Parmar, Jakob Uszkoreit, Llion Jones,
  Aidan~N Gomez, \L~ukasz Kaiser, and Illia Polosukhin.
\newblock Attention is all you need.
\newblock In I.~Guyon, U.~Von Luxburg, S.~Bengio, H.~Wallach, R.~Fergus,
  S.~Vishwanathan, and R.~Garnett, editors, {\em Advances in Neural Information
  Processing Systems}, volume~30. Curran Associates, Inc., 2017.

\bibitem[\protect\citeauthoryear{Wu \bgroup \em et al.\egroup }{2020}]{DBSN}
Xiaohe Wu, Ming Liu, Yue Cao, Dongwei Ren, and Wangmeng Zuo.
\newblock Unpaired learning of deep image denoising.
\newblock In {\em Computer Vision – ECCV 2020: 16th European Conference,
  Glasgow, UK, August 23–28, 2020, Proceedings, Part IV}, page 352–368,
  Berlin, Heidelberg, 2020. Springer-Verlag.

\bibitem[\protect\citeauthoryear{Xu \bgroup \em et al.\egroup
  }{2015}]{Xu_2015_ICCV}
Jun Xu, Lei Zhang, Wangmeng Zuo, David Zhang, and Xiangchu Feng.
\newblock Patch group based nonlocal self-similarity prior learning for image
  denoising.
\newblock In {\em Proceedings of the IEEE International Conference on Computer
  Vision (ICCV)}, December 2015.

\bibitem[\protect\citeauthoryear{Yue \bgroup \em et al.\egroup }{2020}]{DANet}
Zongsheng Yue, Qian Zhao, Lei Zhang, and Deyu Meng.
\newblock Dual adversarial network: Toward real-world noise removal and noise
  generation.
\newblock In Andrea Vedaldi, Horst Bischof, Thomas Brox, and Jan-Michael Frahm,
  editors, {\em Computer Vision -- ECCV 2020}, pages 41--58, Cham, 2020.
  Springer International Publishing.

\bibitem[\protect\citeauthoryear{Zamir \bgroup \em et al.\egroup
  }{2022}]{Restormer}
Syed~Waqas Zamir, Aditya Arora, Salman Khan, Munawar Hayat, Fahad~Shahbaz Khan,
  and Ming-Hsuan Yang.
\newblock Restormer: Efficient transformer for high-resolution image
  restoration.
\newblock In {\em Proceedings of the IEEE/CVF Conference on Computer Vision and
  Pattern Recognition (CVPR)}, pages 5728--5739, June 2022.

\bibitem[\protect\citeauthoryear{Zhang \bgroup \em et al.\egroup
  }{2017}]{DnCNN}
Kai Zhang, Wangmeng Zuo, Yunjin Chen, Deyu Meng, and Lei Zhang.
\newblock Beyond a gaussian denoiser: Residual learning of deep cnn for image
  denoising.
\newblock {\em IEEE Transactions on Image Processing}, 26(7):3142--3155, 2017.

\bibitem[\protect\citeauthoryear{Zhou \bgroup \em et al.\egroup }{2020}]{Zhou}
Yuqian Zhou, Jianbo Jiao, Haibin Huang, Yang Wang, Jue Wang, Honghui Shi, and
  Thomas Huang.
\newblock When awgn-based denoiser meets real noises.
\newblock {\em Proceedings of the AAAI Conference on Artificial Intelligence},
  34(07):13074--13081, Apr. 2020.

\end{thebibliography}

\end{document}